%% file: main.tex
\documentclass[conference]{IEEEtran}
\IEEEoverridecommandlockouts
\usepackage{cite}
\usepackage{amsmath,amssymb,amsfonts}
\usepackage{algorithmic}
\usepackage{graphicx}
\usepackage{textcomp}
\usepackage{xcolor}

\usepackage[absolute,overlay]{textpos}
\usepackage{lipsum}
\usepackage{microtype}
\usepackage{graphicx}
\usepackage{subfigure}
\usepackage{booktabs}
\usepackage{multirow}
\usepackage{enumitem}
\usepackage[most]{tcolorbox}
\usepackage{xcolor}
\usepackage{booktabs}

\usepackage{color}

\usepackage{algorithm}
\usepackage{hyperref}

\newtheorem{theorem}{Theorem}

\def\BibTeX{{\rm B\kern-.05em{\sc i\kern-.025em b}\kern-.08em
    T\kern-.1667em\lower.7ex\hbox{E}\kern-.125emX}}
\begin{document}

\title{Free-T2M: Robust Text-to-Motion Generation for Humanoid Robots via Frequency-Domain Consistency
}

\author{
\IEEEauthorblockN{1\textsuperscript{st} Wenshuo Chen\textsuperscript{*}}
\IEEEauthorblockA{
\textit{$DI^2$ Lab, HKUST(GZ)}\\
Guangzhou, China \\
wenshuochen@hkust-gz.edu.cn
}
\and
\IEEEauthorblockN{2\textsuperscript{nd} Haozhe Jia\textsuperscript{*}}
\IEEEauthorblockA{
\textit{$DI^2$ Lab, HKUST(GZ)}\\
\textit{Shandong University}\\
Qingdao, China \\
haozhejia@hkust-gz.edu.cn
}
\and
\IEEEauthorblockN{3\textsuperscript{rd} Songning Lai}
\IEEEauthorblockA{
\textit{$DI^2$ Lab, HKUST(GZ)}\\
Guangzhou, China \\
songninglai@hkust-gz.edu.cn
}
\and
\IEEEauthorblockN{4\textsuperscript{th} Lei Wang}
\IEEEauthorblockA{
\textit{Griffith University, Brisbane, Queensland, Australia}\\
\textit{Data61/CSIRO, Canberra, ACT, Australia}\\
l.wang4@griffith.edu.au
}
\and
\IEEEauthorblockN{5\textsuperscript{th} Yuqi Lin}
\IEEEauthorblockA{
\textit{$DI^2$ Lab, HKUST(GZ)}\\
Guangzhou, China \\
yuqilin@hkust-gz.edu.cn
}
\and
\IEEEauthorblockN{6\textsuperscript{th} Hongru Xiao}
\IEEEauthorblockA{
\textit{Tongji University}\\
Shanghai, China \\
hongruxiao@tongji.edu.cn
}
\and
\IEEEauthorblockN{7\textsuperscript{th} Lijie Hu}
\IEEEauthorblockA{
\textit{MBZUAI}\\
Abu Dhabi, United Arab Emirates \\
lijie.hu@mbzuai.ac.ae
}
\and
\IEEEauthorblockN{8\textsuperscript{th} Yutao Yue}
\IEEEauthorblockA{
\textit{$DI^2$ Lab, HKUST(GZ)}\\
Guangzhou, China \\
yutaoyue@hkust-gz.edu.cn
}
\thanks{* Equal contribution. Corresponding author: yutaoyue@hkust-gz.edu.cn}
}



\maketitle

\begin{figure*}[!ht]
    \centering
    \includegraphics[width=\linewidth]{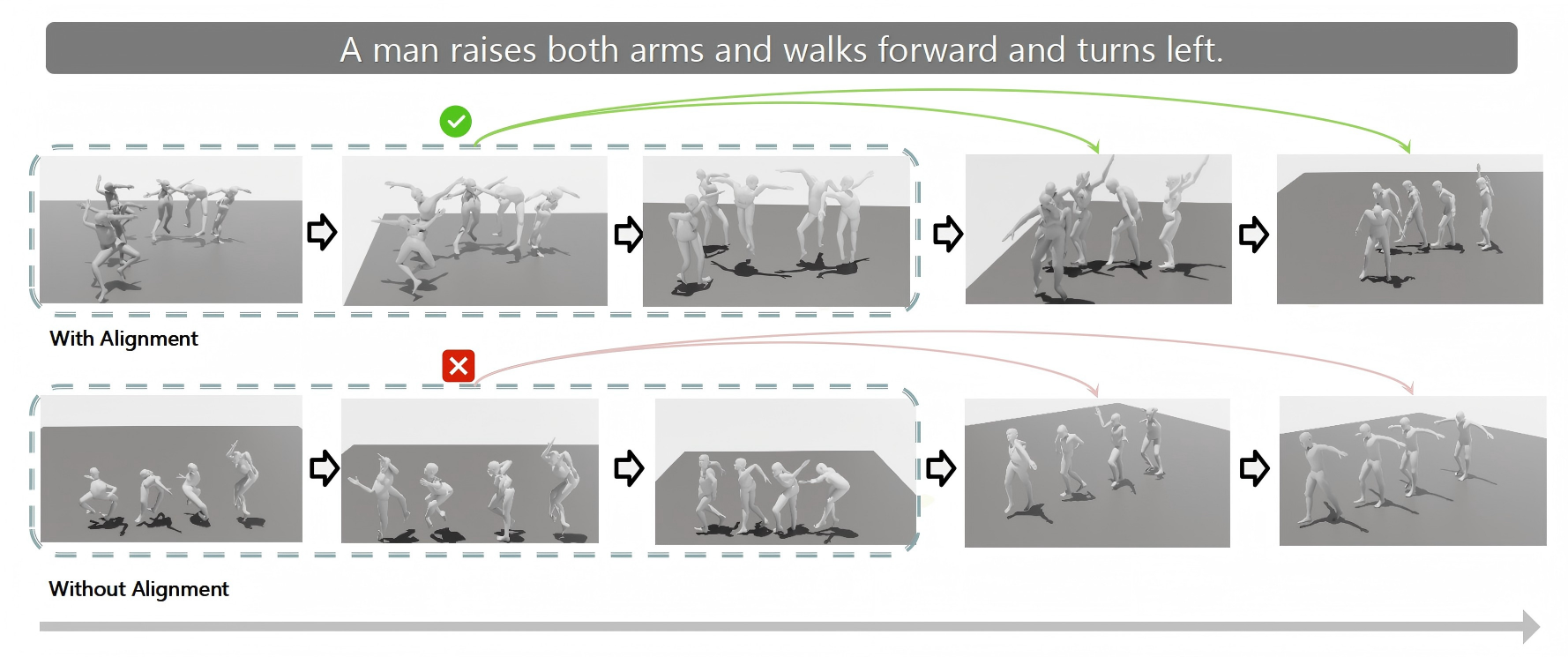}
    \caption{T2M Denoising as Hierarchical Motion Planning: Early stages serve as semantic planning for low-frequency recovery, later stages refine high-frequency details. As shown in the lower part of the figure, failures in low-frequency semantics lead to cascading errors, while poor high-frequency refinement yields incoherent motions. Our frequency-adaptive alignment enhances planning, improving semantic fidelity and naturalness.}
    \label{fig:intro}
    \vspace{-0.2cm}
\end{figure*}

\input{sec/0_abstract}


\input{sec/1_intro}

\input{sec/2_related_work}
\input{sec/3_method}

\input{sec/4_experiment}

\input{sec/5_conclusion}

\bibliography{example_paper}
\bibliographystyle{ieeetr}



\end{document}

%% file: sec/0_abstract.tex
\begin{abstract}
Enabling humanoid robots to synthesize complex, physically coherent motions from natural language commands is a cornerstone of autonomous robotics and human-robot interaction. While diffusion models have shown promise in this text-to-motion (T2M) task, they often generate semantically flawed or unstable motions, limiting their applicability to real-world robots. This paper reframes the T2M problem from a frequency-domain perspective, revealing that the generative process mirrors a hierarchical control paradigm. We identify two critical phases: a \textbf{semantic planning stage}, where low-frequency components establish the global motion trajectory, and a \textbf{fine-grained execution stage}, where high-frequency details refine the movement. To address the distinct challenges of each phase, we introduce \textbf{\underline{Fre}}quency \textbf{\underline{e}}nhanced \textbf{\underline{t}}ext-\textbf{\underline{to}}-\textbf{\underline{m}}otion (\textbf{Free-T2M}), a framework incorporating stage-specific frequency-domain consistency alignment. We design a frequency-domain temporal-adaptive module to modulate the alignment effects of different frequency bands. These designs enforce robustness in the foundational semantic plan and enhance the accuracy of detailed execution. Extensive experiments show our method dramatically improves motion quality and semantic correctness. Notably, when applied to the StableMoFusion baseline, Free-T2M reduces the FID from \textbf{0.152} to \textbf{0.060}, establishing a new state-of-the-art within diffusion architectures. These findings underscore the critical role of frequency-domain insights for generating robust and reliable motions, paving the way for more intuitive natural language control of robots. 
\end{abstract}

%% file: sec/1_intro.tex
\section{Introduction}
\label{sec:intro}
%
The ability to interpret high-level natural language commands and translate them into complex, physically coherent motion sequences is a fundamental goal for the next generation of autonomous systems, particularly humanoid robots \cite{MDM, SATO, jiang2024harmonwholebodymotiongeneration, luo2024universalhumanoidmotionrepresentations}. Such a capability would unlock intuitive human-robot interaction in diverse applications, from assistive care and collaborative manufacturing to domestic services. While significant progress has been made in robot control and motion planning, generating diverse and contextually appropriate motions directly from unstructured text remains a formidable challenge. Diffusion models, with their remarkable success in other generative domains \cite{DDPM, stablevideodiffusionscaling, ning2025dctdiffintriguingpropertiesimage}, have recently emerged as a powerful paradigm for this text-to-motion (T2M) task \cite{MLD, stablemofusion, dai2024motionlcmrealtimecontrollablemotion}, promising to bridge the gap between human language and robot execution.

However, despite their potential, current T2M models designed for motion synthesis often exhibit a critical brittleness that hinders their deployment on physical robotic platforms \cite{MDM,stablemofusion,T2M-GPT}. These models, which primarily focus on temporal sequence modeling, can produce motions that are physically implausible, lack semantic coherence, or fail to capture the natural rhythm of human actions \cite{humanmotion3d, motionx}. A slight perturbation in the input text can lead to drastically different and often incorrect robot behaviors, a reliability issue highlighted by prior work \cite{SATO}. This lack of robustness stems from an incomplete understanding of the motion generation process, as existing methods often adapt general-purpose diffusion techniques without tailoring them to the unique structure of motion data \cite{SATO, miao2024autonomousllmenhancedadversarialattack}.

Our work addresses this gap by investigating the denoising mechanism of diffusion models from a frequency-domain perspective, which offers a more interpretable and effective solution. As illustrated in Figure \ref{fig:intro}, we observe that the denoising process inherently mimics a hierarchical planning and control strategy common in robotics. During the early stages, the model primarily recovers low-frequency components, corresponding to the broad, semantic aspects of the motion (e.g., the overall trajectory of the torso). This can be viewed as a \textbf{task-level semantic planning stage}. Subsequently, in later stages, the model refines this plan by synthesizing high-frequency components, which represent fine-grained details of limb movement. This corresponds to a \textbf{low-level motion execution stage}. We posit that the root cause of catastrophic failures is an erroneous semantic plan in the early stages; once the high-level trajectory is wrong, no amount of subsequent refinement can produce a correct motion.

Based on this insight, we first apply the Discrete Cosine Transform (DCT) as a frequency-domain conversion method. Initially, we attempted to manually specify the alignment frequency bands, which led to unstable training and made it difficult to determine which frequency bands should be aligned at different timesteps. Inspired by \cite{Chen2025ANTAN}, we design a \textbf{f}requency-domain \textbf{t}emporal-\textbf{a}daptive module (FTA) that adaptively adjusts the frequency band alignment at different timesteps during training and discards this information during inference. Experiments demonstrate that our method is simple yet effective. On HumanML3D, we achieve significant improvements over the baseline in both FID (\textbf{0.060} vs 0.152) and R-Precision (\textbf{0.555} vs 0.546), confirming that our approach can substantially enhance both the accuracy and semantic consistency of generated motions.


Our main contributions, framed for the robotics community, are as follows:
\begin{itemize}[topsep=0pt,itemsep=2pt,parsep=0pt,partopsep=0pt] 
    \item We introduce a novel perspective for analyzing text-conditioned motion generation, framing the diffusion denoising process as a hierarchical paradigm of semantic planning and fine-grained execution.
    \item We propose stage-specific adaptive frequency-domain consistency alignment that significantly enhances the robustness of the high-level motion plan and the precision of low-level execution, leading to more physically plausible and semantically correct motions.
    \item Through extensive experiments, we demonstrate that our plug-and-play approach achieves state-of-the-art performance on standard benchmarks without increasing model complexity, offering a practical solution for advancing natural language control in robotic systems.
\end{itemize}

%% file: sec/2_related_work.tex
\section{Related Work}

\subsection{From High-Level Commands to Robot Motion}

Translating high-level, often abstract, human commands into executable trajectories for robots is a long-standing challenge in robotics \cite{firoozi2025foundation,chen2024autotamp}. Classical approaches often rely on structured hierarchical frameworks, such as Task and Motion Planning (TAMP) \cite{kaelbling2011hierarchical,vu2024coast}. TAMP systems decompose a problem into a high-level symbolic planner that determines a sequence of actions (e.g., ``pick(object)'', ``move(location)'') and a low-level motion planner that computes collision-free paths to realize those actions. While powerful, these methods typically require formal symbolic representations of the task and environment, limiting their ability to handle the ambiguity and richness of natural language \cite{gatt2018survey,cohen2024survey}. Another dominant paradigm is Learning from Demonstration (LfD) \cite{schaal1996learning,atkeson1997robot,brys2015reinforcement,dubois2023alpacafarm}, where robots learn motion policies by imitating human examples. LfD excels at capturing the nuances of human movement but often struggles with generalizing to new tasks or instructions not seen during training \cite{barreiros2025motor,huang2024surface}. Our work draws inspiration from these hierarchical concepts but tackles the problem from a fundamentally different, end-to-end generative modeling perspective, aiming to directly map unstructured text to complex motion sequences without explicit symbolic reasoning.

\subsection{Text Conditioned Human Motion Diffusion Generation} 

Recent advancements in text-conditioned human motion generation have been driven by diffusion models \cite{MDM,remodiffuse,stablemofusion,motiondiffuse,MLD,dai2024motionlcmrealtimecontrollablemotion}, which have shown remarkable potential in modeling the complex relationships between textual inputs and motion sequences. Prominent works include MotionDiffuse \cite{motiondiffuse}, which leverages cross-attention for text integration; MDM \cite{MDM}, which explores diverse denoising networks such as Transformer and GRU; and PhysDiff \cite{physdiff}, which incorporates physical constraints to enhance realism. While ReMoDiffuse \cite{remodiffuse} improves performance through retrieval mechanisms, MotionLCM \cite{dai2024motionlcmrealtimecontrollablemotion} achieves real-time, controllable generation via a latent consistency model. Despite these advances, existing methods often struggle with capturing both low-frequency semantic structures and high-frequency motion details, which are critical for generating coherent and realistic human motions.

Moreover, diffusion models in motion generation often lack tailored designs for the intrinsic properties of motion data. For example, the denoising process may fail to prioritize low-frequency components that define the overall motion trajectory, leading to semantic inconsistencies in the generated outputs \cite{amass,MDM}. Furthermore, the reliance on generic architectures \cite{li2024infinitemotionextendedmotion,hu2024efficienttextdrivenmotiongeneration} limits their ability to address domain-specific challenges, such as robust handling of complex text prompts and preserving temporal coherence.

%% file: sec/3_method.tex
\section{Method}

\begin{figure}[t]
    \centering
    \includegraphics[width=\linewidth]{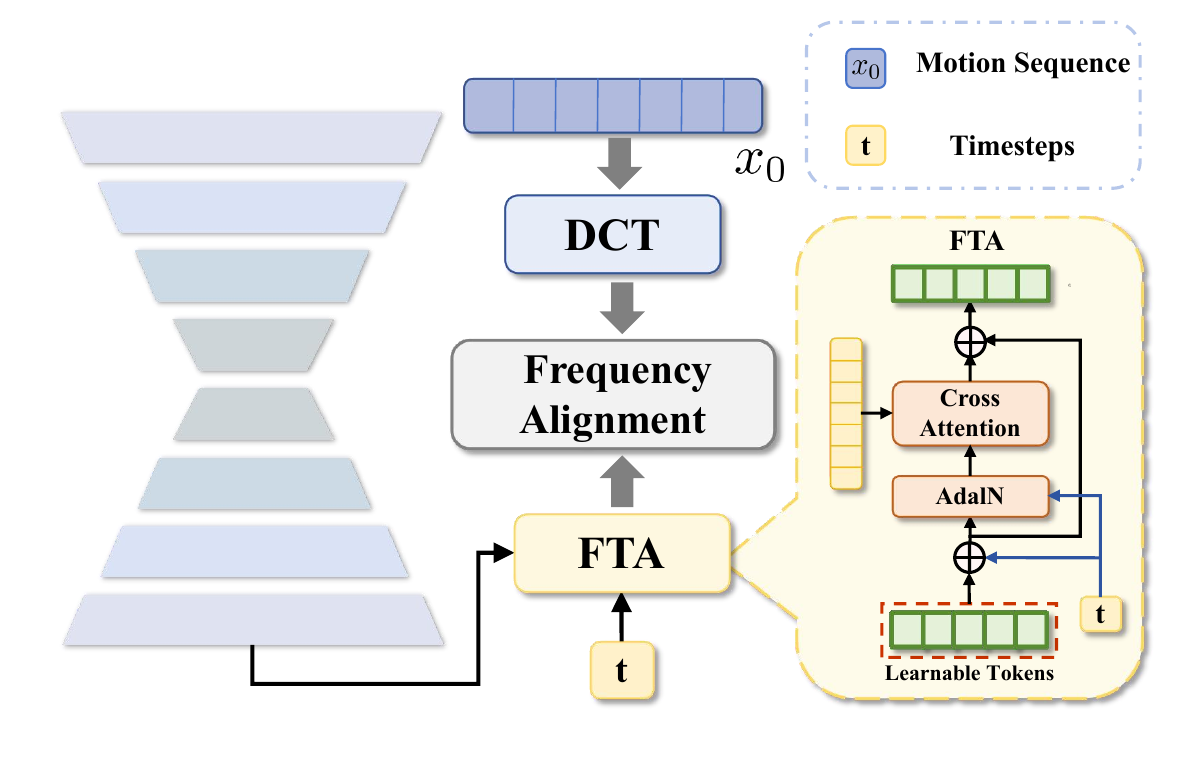}
    \caption{\textbf{Overview of the Free-T2M Framework for Robotic Motion Synthesis.} Free-T2M employs a UNet architecture featuring two core components. frequency-domain temporal-adaptive module (FTA) conditions the temporal attention on the diffusion timestep $t$, enabling an adaptive focus that shifts from coarse, task-level planning (low frequencies) in early denoising stages to fine-grained motor control (high frequencies) in later stages. An auxiliary head, supervised in the full Discrete Cosine Transform (DCT) space, ensures multi-scale temporal consistency, which is critical for generating physically plausible and executable robotic motions.}
    \label{fig:Free-T2M}
\end{figure}

\subsection{Preliminaries}

The diffusion model generation process can be divided into three stages: a \textbf{forward process}, which progressively diffuses noise into the data sample; a \textbf{reverse process}, where a network is optimized to remove added noise and recover the original signal; and an \textbf{inference process}, which employs the trained network to iteratively denoise noisy samples.

In the forward diffusion process, noise is incrementally introduced to the data $x_0$ to yield $x_t$. In the frequency domain, this can be expressed as:
\begin{equation}
X_t(f) = \sqrt{\bar{\alpha}_t} X_0(f) + \sqrt{1 - \bar{\alpha}_t} \epsilon(f),
\end{equation}
where $\bar{\alpha}_t$ and $\epsilon(f)$ represent the preserved signal strength and Gaussian noise in the frequency domain, respectively. Notably, during this process, the high-frequency components of the motion data are disrupted first, followed by the low-frequency components.

\begin{theorem}
In diffusion models, low-frequency components are recovered earlier than high-frequency components during the reverse denoising process \cite{boostingdiffusionmodelsmoving}. For a motion signal, the signal-to-noise ratio (SNR) for higher frequencies $\omega_H$ decreases more rapidly than for lower frequencies $\omega_L$:
\begin{equation}
    \text{SNR}(\omega_H) < \text{SNR}(\omega_L), \quad \forall \, \omega_H > \omega_L.
\end{equation}
As a result, low-frequency components are restored first, providing a coarse foundation for the subsequent recovery of high-frequency details.
\end{theorem}

In the reverse process, the model starts from pure noise $x_T$ and iteratively reconstructs the original motion. Initially, only low-frequency information, such as broad motion trajectories, is recovered. As the denoising progresses (as $t$ decreases), the model gradually restores high-frequency details and finer motion features. This hierarchical, coarse-to-fine reconstruction process is a fundamental property of diffusion models.

\subsection{From Hierarchical Denoising to Robotic Motion Planning}

This hierarchical recovery process provides a powerful analogy for robotic motion planning, which is also inherently hierarchical. To bridge the gap between a high-level natural language command and a low-level, executable trajectory, a robot must first establish a coarse, task-level plan and then refine it with fine-grained motor control. We map these concepts to the frequency domain:
\begin{itemize}
    \item \textbf{Low-frequency components} correspond to task-level semantics: the robot's overall trajectory, global posture changes, and the strategic path to fulfill the command (e.g., ``walk to the door'' ).
    \item \textbf{High-frequency components} correspond to fine motor coordination: precise joint articulations, end-effector control, and subtle balance adjustments required for physical plausibility and stable execution.
\end{itemize}

Our framework, Free-T2M, is designed to explicitly leverage this correspondence. It uses the diffusion timestep $t$ as a signal to control the level of abstraction it focuses on during generation, ensuring that task-level plans are established before fine details are synthesized. This is achieved through two key technical contributions: a Time-Aware Self-Temporal Attention mechanism and a DCT-space auxiliary supervision signal.

\subsection{Time-Aware Self-Temporal Attention for Adaptive Motion Planning}

The key insight driving our approach is that different stages of the reverse diffusion process correspond to different levels of motion abstraction in robotic control. Early denoising steps (large $t$) should focus on establishing task-level motion plans, while later steps (small $t$) should refine fine motor coordination. To achieve this adaptive behavior, we introduce a Time-Aware Self-Temporal Attention (FTA) mechanism.

Given an intermediate motion representation $X \in \mathbb{R}^{B \times T \times D}$ from the UNet backbone, the FTA module first computes a timestep embedding $e_t$:
\begin{equation}
    e_t = f_{\text{MLP}}(\text{PE}(t)),
\end{equation}
where $\text{PE}(\cdot)$ denotes sinusoidal positional encoding. The core innovation lies in using $e_t$ to modulate the attention computation through Adaptive Layer Normalization, allowing the model to dynamically adjust its temporal focus based on the denoising stage:
\begin{equation}
    \tilde{X} = \text{LayerNorm}(X) \cdot (1 + \gamma(e_t)) + \beta(e_t),
\end{equation}
where $\gamma(\cdot)$ and $\beta(\cdot)$ are learned linear projections of the timestep embedding. The attention mechanism then operates on the modulated representations:
\begin{align}
    Q &= W_q \tilde{X}, \\
    K &= W_k \tilde{X}, \\
    V &= W_v \tilde{X},
\end{align}
where $W_q$, $W_k$, and $W_v$ are learned projection matrices. The multi-head attention output is computed as:
\begin{equation}
    \text{Attention}(Q, K, V) = \text{softmax}\left(\frac{QK^T}{\sqrt{d_k}}\right)V,
\end{equation}
and the final FTA output incorporates a residual connection:
\begin{equation}
    \text{FTA}(X, t) = X + \text{Attention}(Q, K, V).
\end{equation}

This time-conditioned attention enables the model to exhibit distinct behaviors across denoising stages:
\begin{itemize}
    \item \textbf{Early stages} (large $t$): The attention mechanism emphasizes long-range temporal dependencies, facilitating the establishment of coherent, low-frequency task-level trajectories.
    \item \textbf{Later stages} (small $t$): The focus shifts to local temporal relationships, enabling the refinement of high-frequency details for precise joint coordination and stable execution.
\end{itemize}

\subsection{DCT-Space Auxiliary Supervision for Multi-Scale Temporal Consistency}

While FTA provides adaptive temporal modeling, we further stabilize the learning process through explicit frequency-domain supervision. Unlike methods that focus only on low frequencies, our approach supervises the model on the entire frequency spectrum. This is crucial for robotics, as the relationship between all frequency components must be preserved to generate physically executable motions without artifacts like jitter or foot sliding.

We introduce an auxiliary prediction branch that operates in the Discrete Cosine Transform (DCT) space. For a given diffusion target $Y$ (e.g., the clean motion or noise), we compute its frequency-domain representation $\hat{Y} = \text{DCT}(Y)$ using an orthonormal DCT. The auxiliary branch, conditioned on the FTA output, produces a prediction $\hat{X}^{\text{dct}}$ that is supervised directly against this full-spectrum target. To handle variable-length motion sequences, we use a masked loss:
\begin{equation}
    \mathcal{L}_{\text{dct}} = \frac{1}{B}\sum_{b=1}^{B} \frac{\sum_{t=1}^{T} M_{b,t} \|\hat{X}^{\text{dct}}_{b,t,:} - \hat{Y}_{b,t,:}\|_2^2}{\sum_{t=1}^{T} M_{b,t}},
\end{equation}
where $M \in \{0,1\}^{B \times T}$ is a binary mask for valid frames. This loss encourages the generation of motions with realistic temporal dynamics across all frequencies.

\subsection{Training Objective}

The complete training objective for Free-T2M combines the standard time-domain diffusion loss with our DCT-space auxiliary supervision:
\begin{equation}
    \mathcal{L}_{\text{total}} = \mathcal{L}_{\text{diffusion}} + \lambda_{\text{dct}} \mathcal{L}_{\text{dct}},
\end{equation}
where $\mathcal{L}_{\text{diffusion}}$ is the masked mean squared error in the time domain, and $\lambda_{\text{dct}}$ is a weighting coefficient. This synergistic approach uses FTA for adaptive, hierarchical planning and full-spectrum DCT supervision to ensure the physical plausibility and temporal coherence of the final motion, bridging the gap between high-level language commands and low-level robotic execution.

%% file: sec/4_experiment.tex
\section{Experiment}

\begin{figure*}[th]
    \centering
    \includegraphics[width=\linewidth]{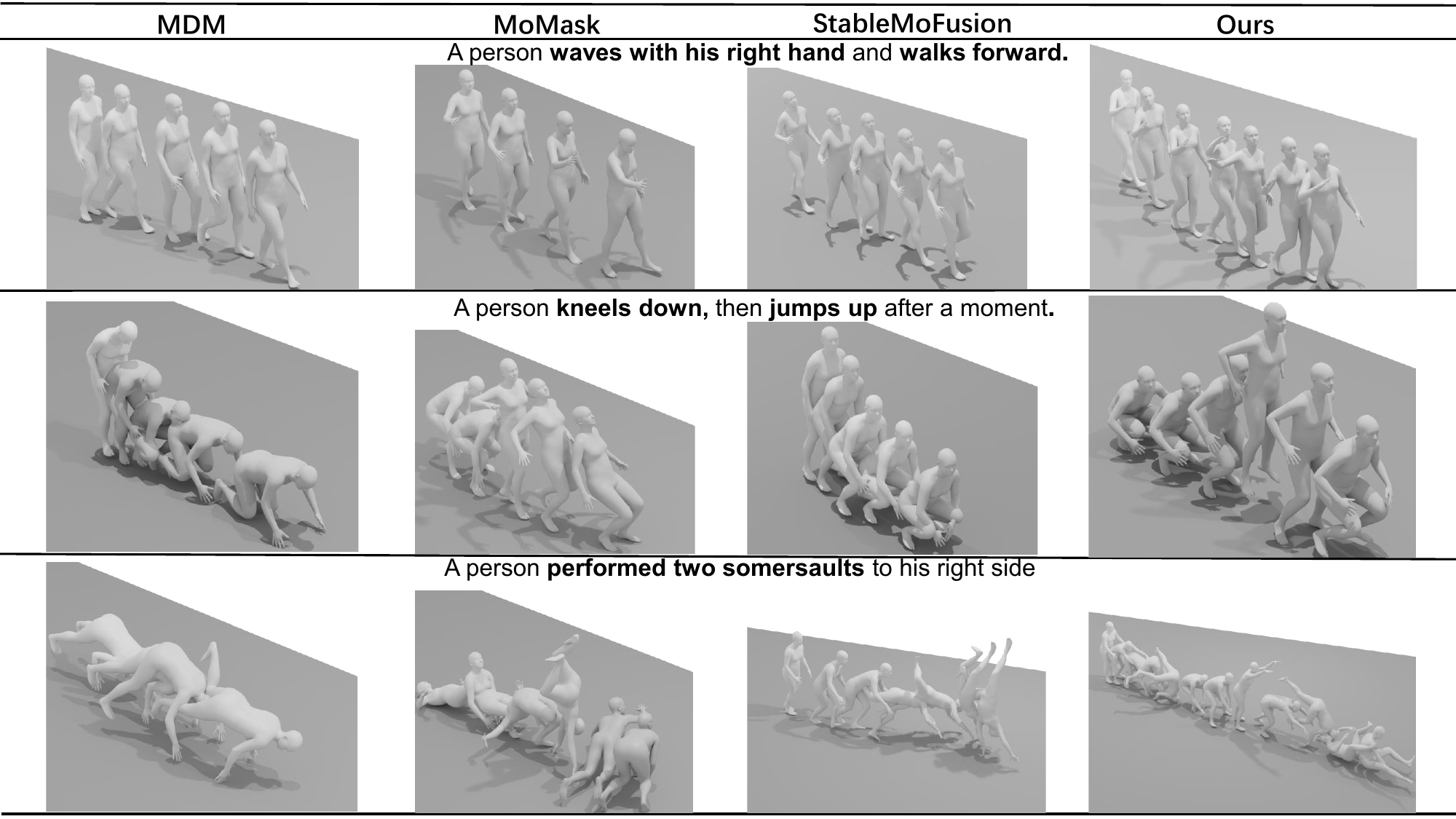}
    \caption{Visualization results. The white arrows represent motion trajectories. In several examples, the baseline model produces errors in low-frequency information, leading to incorrect trajectories and ultimately affecting the accuracy of the motion. Our model demonstrates superior semantic consistency and finer-grained local actions compared to the baseline.}
    \label{fig:visualization}
\end{figure*}

\input{tab/humanml}

We evaluate our approach on two popular motion-language benchmarks: HumanML3D \cite{Guo_2022_CVPR} and KIT-ML \cite{KIT}. The HumanML3D dataset comprises 14,616 motions sourced from AMASS \cite{AMASS:ICCV:2019} and HumanAct12 \cite{guo2020action2motion}, each paired with three textual descriptions, resulting in a total of 44,970 descriptions. This dataset encompasses a wide range of actions, such as walking, exercising, and dancing. The KIT-ML dataset, on the other hand, includes 3,911 motions and 6,278 text descriptions, serving as a smaller-scale evaluation benchmark. For both datasets, we employ the pose representation used in T2M-GPT. Motions are augmented through mirroring and split into training, validation, and testing sets with a ratio of 0.8 : 0.15 : 0.05. \cite{T2M-GPT, guo2023momaskgenerativemaskedmodeling}

\noindent \textbf{Evaluation Metrics.}
In addition to the commonly utilized metrics such as Frechet Inception Distance (FID), R-Precision, Multimodal Distance (MM-Dist), and Diversity, which are employed by T2M-GPT \cite{T2M-GPT}. Furthermore, human evaluation is employed to obtain accuracy and human preference results for the outputs generated by the model.
\begin{itemize}[noitemsep,leftmargin=*]

    \item \textbf{Frechet Inception Distance \cite{NIPS2017_8a1d6947} (FID):} We can evaluate the overall motion quality by measuring the distributional difference between the high-level features of the motions.
    \item \textbf{R-Precision:} We rank Euclidean distances between a given motion sequence and 32 text descriptions (1 ground-truth and 31 randomly selected mismatched descriptions). We report Top-1, Top-2, and Top-3 accuracy of motion-to-text retrieval.
    
    \item \textbf{Diversity:} From a set of motions, we randomly sample 300 pairs and compute the average Euclidean distances between them to measure motion diversity.
    
    \item \textbf{Human Evaluation:} We conducted evaluations of each model's generated results in the form of a Google Form. We collected user ratings on model prediction, which encompassed both the quality and correctness of the generated motions. Additionally, we analyzed user preferences for model prediction.
\end{itemize}

\subsection{Experimental Setup}
We adopt model architecture settings similar to  StableMoFusion.
Our model uses a 1D UNet with base channels of 512, channel multipliers $[2,2,2,2]$, and a Transformer text encoder of dimension 256. The FTA module applies timestep embeddings via a two-layer MLP and AdaLayerNorm with a single attention head. HumanML3D and KIT-ML datasets use their respective pose feature dimensions ($d_{\text{model}}=263$ and $251$). We apply full-spectrum DCT supervision with weight $\lambda_{\text{dct}}=0.2$, and low-frequency supervision ($k=25$) for ablations. Training uses AdamW (lr $2\times10^{-4}$, weight decay $10^{-2}$), batch size 64, EMA 0.9999, linear $\beta$ schedule with $T=1000$, and classifier-free guidance (masking 0.1). Experiments run on a single RTX 4090 GPU with optional multi-GPU support.

\input{tab/ablation_Loss}
\subsection{Comparison with Baseline}
\subsubsection{Quantitative comparisons.}
We evaluated our model on the HumanML3D and KIT datasets, comparing its performance to baseline models (Table \ref{tab:combined_results}. For StableMoFusion, our approach achieved significant improvements in both accuracy (Top-2 \textbf{0.742 $\to$ 0.752}) and precision (FID \textbf{0.152 $\to$ 0.060}) in HumanML3D dataset. Similarly, on KIT-ML, our method achieved notable improvements in accuracy.


Additionally, we compared our method with ReModiffuse \cite{remodiffuse}, MotionLCM \cite{dai2024motionlcmrealtimecontrollablemotion}, MotionDiffuse \cite{motiondiffuse}, MoMask \cite{guo2023momaskgenerativemaskedmodeling} and T2M-GPT \cite{T2M-GPT}. The results demonstrate that our Free-T2M achieves state-of-the-art (SOTA) performance in R-Precision and MM-Dist. This indicates that our method enables the model’s predictions to better align with semantic information. Meanwhile, the substantial improvement in FID further demonstrates that the frequency-domain alignment mechanism helps the model better capture the underlying data distribution.

\subsubsection{Qualitative comparison.}

Figure \ref{fig:visualization} presents visual comparisons with MDM, StableMoFusion, and MoMask under three prompts. For ``A person waves with his right hand and walks forward, '' MDM and StableMoFusion only ``walk'', while MoMask fails to wave properly. Our model performs the full
action accurately. For ``A person kneels down onto the floor then waits and raises his hands up,'' MDM and StableMoFusion miss kneeling, and MoMask fails to jump up; our method captures both
with correct timing. These two examples require the model to accurately and coherently generate multiple motions in a temporal sequence. However, previous models often fail to adhere to semantic information, producing motions that are missing, erroneous, or unrealistic. In contrast, our model demonstrates significantly superior temporal-semantic precision compared to other models. For the complex prompt with requirements on repetition count and spatial position, "A person performed two somersaults to his right side", the baseline models performed extremely poorly, all generating erroneous results, while our model executed it with high fidelity. These results demonstrate superior semantic understanding, temporal consistency, and detail accuracy.

\begin{figure}[htbp]
    \centering
    \includegraphics[width=\linewidth]{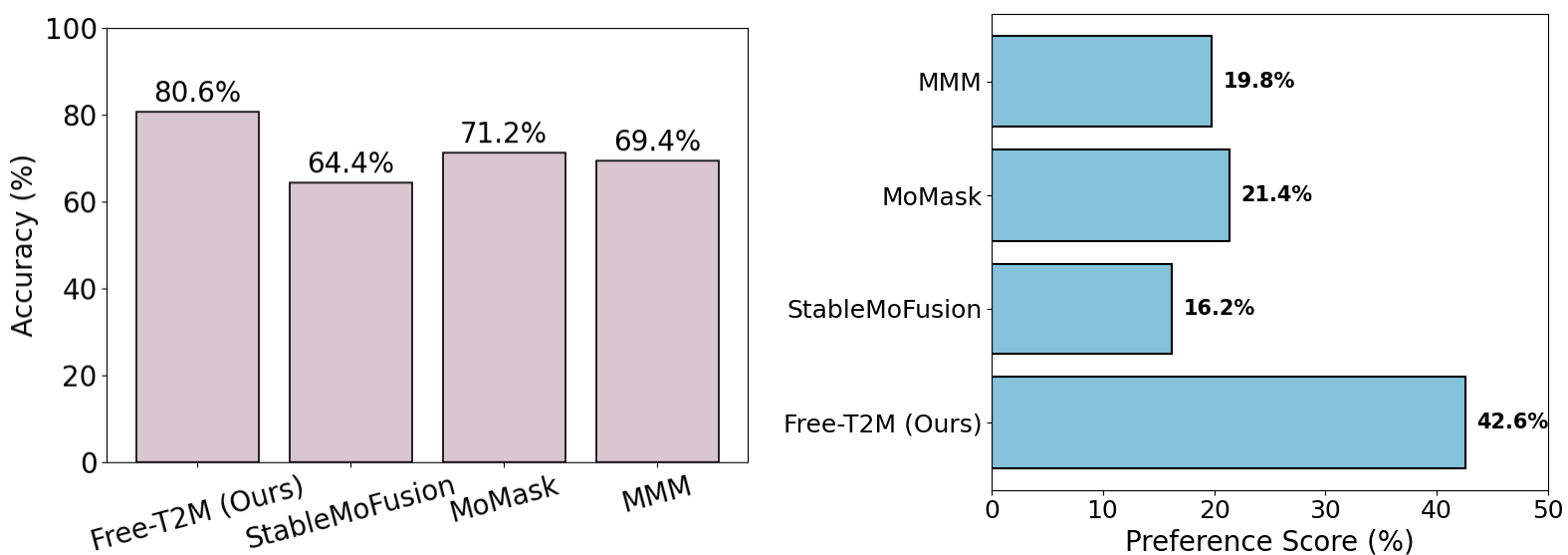}
    \caption{Comparative performance of ANT (StableMoFusion) versus other baseline methods based on human evaluation. The left panel shows the accuracy (\%) obtained during manual assessments. The right panel presents the preference score, computed as the proportion of times each method’s generated motion was selected as the most preferred among all four methods in each evaluation.}
    \label{fig:human eval}
\end{figure}

\subsection{Human Evaluation}
\label{sec:human evaluation}
We randomly sampled 100 text entries from the HumanML3D Test dataset, and under the same random seed and other parameter settings, we used Free-T2M (ours), StableMoFusion (baseline), MoMask and MMM to generate corresponding motions for visualization. Five different evaluators anonymously compared the visualized outputs with the ground truth (GT) in a pairwise manner for manual evaluation. Evaluators were asked to assess the correctness of the motions and rank the models based on naturalness, accuracy, and completeness to select the better-performing model. The results are shown in the Figure \ref{fig:human eval}. Our model demonstrated a significant improvement in accuracy, with increases of \textbf{16.2}\% over StableMoFusion (baseline). Additionally, a greater number of users preferred the results generated by our model, with an advantage of \textbf{21.2}\% over MoMask and \textbf{26.4}\% over StableMoFusion. Overall, our approach better preserves the consistency of generated low-frequency semantic content with the condition while significantly enhancing high-frequency motion details, thereby further improving the coherence and naturalness of the motions. These results highlight the effectiveness of our method.

\begin{figure}[t]
    \centering
    \includegraphics[width=\linewidth]{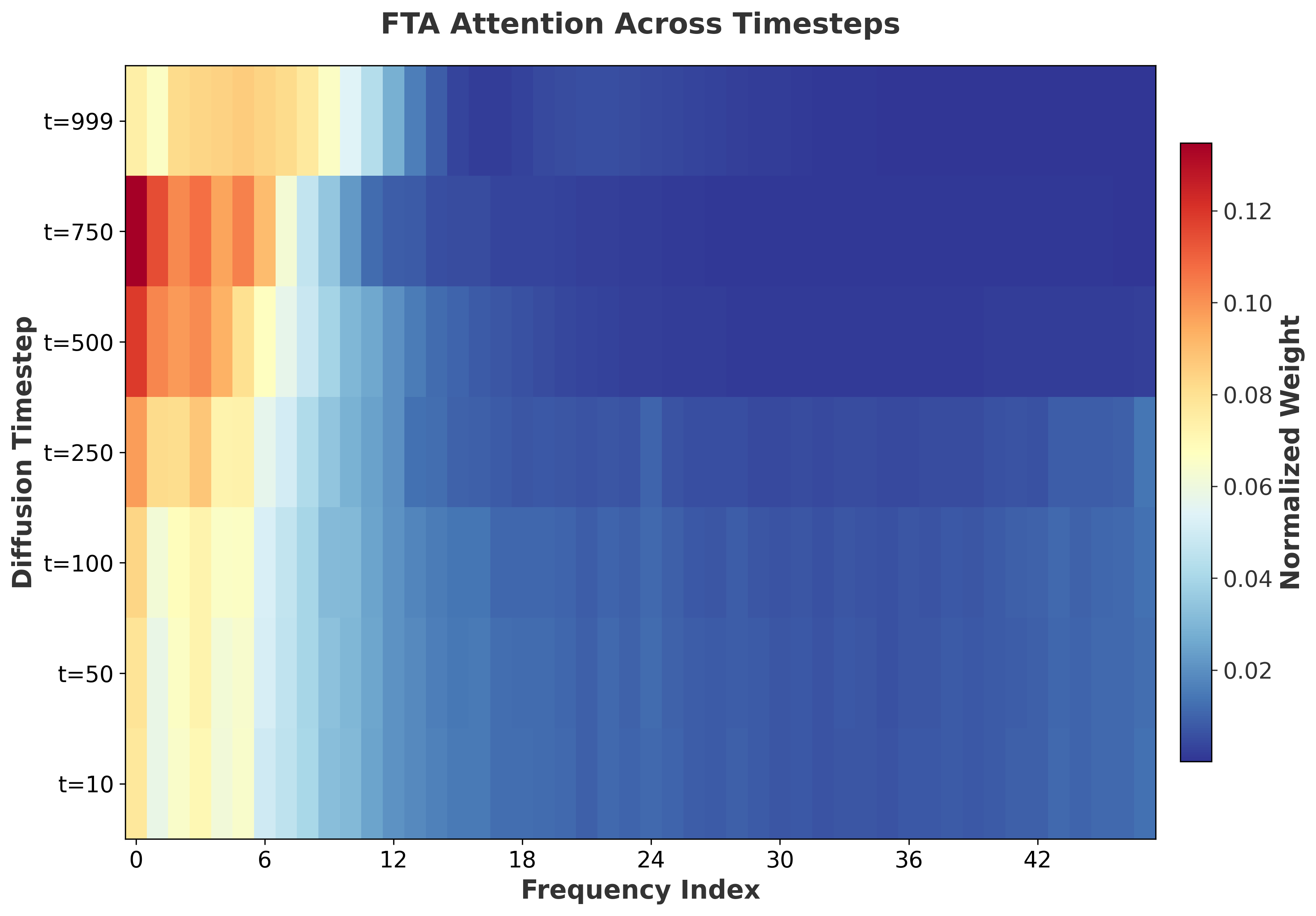}
    \caption{\textbf{FTA attention across timesteps.} Row-normalized heatmap of frequency weights computed by applying a DCT to UNet outputs (averaged over joints) with log-compression and re-binned frequency axis. Early denoising steps (large $t$) concentrate on low-frequency bands (left); as $t$ decreases, energy shifts toward mid/high frequencies (right), evidencing the coarse-to-fine generation behavior and validating the time-adaptive design of FTA and our frequency-alignment objective.}
    \label{fig:fta attention}
\end{figure}


\subsection{Ablation Study}
In Table \ref{table:ablation_results}, We conducted comprehensive ablation studies to validate the effectiveness of our proposed method. In our ablation studies, we investigate the role of the Frequency-Temporal Alignment (FTA) module as well as the choice of frequency bands to be aligned. Specifically, we experiment with low-frequency ($f_l$), high-frequency ($f_h$), and full-frequency ($f_{full}$) bands. When FTA is not applied, we use a simple MLP to project different frequency bands into the feature space for alignment. Several key observations can be drawn. First, regardless of which frequency band is aligned ($f_{full}, f_l, f_h$), incorporating FTA consistently outperforms direct MLP mapping in terms of both FID and R-Precision. This highlights the importance of time-dependent features $t$ in guiding the alignment between frequency and feature spaces. Second, when using FTA, aligning the full frequency band yields the best results. This is because FTA adaptively selects appropriate frequency bands across timesteps—favoring low-frequency alignment in the early stages to better capture semantic planning, and shifting toward high-frequency alignment later to enrich fine-grained motion details. Third, without FTA, low-frequency alignment achieves the best performance. This can be attributed to the low-frequency dominance of motion, where aligning only low-frequency signals enables the model to learn a distribution closer to that of real data. Overall, our findings demonstrate that the combination of FTA and full-frequency alignment provides the most effective and interpretable solution.

\subsection{FTA Attention Analysis}
To examine how the Frequency-domain Temporal-Adaptive (FTA) module allocates attention across temporal frequencies along the reverse diffusion trajectory, we compute a frequency spectrum at several timesteps $t$ by applying a DCT to the UNet outputs, averaging over joints/channels, log-compressing, row-normalizing, and re-binning the frequency axis. The resulting timestep–frequency heatmap (see Fig.~\ref{fig:fta attention}) reveals a clear coarse-to-fine evolution: at large $t$ the mass concentrates on low frequencies, while as $t$ decreases the energy progressively shifts toward mid/high frequencies, indicating that global, task-level structure is prioritized early and fine motor details are refined later. This behavior is consistent with the SNR-based analysis in the Preliminaries (low-frequency components are reconstructed earlier than high-frequency ones) and provides model-internal, frequency-domain evidence for our frequency-alignment objective and the coarse-to-fine denoising insight of Free-T2M.

\subsection{Robustness Analysis}

Following SATO \cite{SATO}, robustness is a critical challenge for T2M models. To address this, we utilized the SATO perturbed text dataset, which involves synonym replacements in the text while preserving its length and meaning. This dataset simulates diverse expressions of the same intent by different users interacting with the model. We tested Free-T2M and StableMoFusion respectively on perturbed texts. The experimental results demonstrate that our method outperforms baseline models on both FID (\textbf{0.168} vs 0.202) and Top-3 (\textbf{0.706} vs 0.660)under text perturbations across two models. This indicates that our approach enhances the model's robustness to some extent, aligning with the motivation behind our method. By incorporating the frequency-domain alignment, the model becomes more stable in capturing low-frequency semantic information, leading to improved accuracy. With accurate low-frequency information in place, the high-frequency information, which depends on the low-frequency context, is better modeled, resulting in more precise generation of details such as hand and leg movements.

%% file: tab/humanml.tex
\begin{table*}[t]
\centering
\resizebox{\textwidth}{!}{%
\begin{tabular}{llccccccc}
\toprule
\textbf{Method} & \textbf{Venue} & \textbf{FID $\downarrow$}  &\multicolumn{3}{c}{\textbf{R-Precision $\uparrow$}} & \textbf{Diversity $\rightarrow$}  & \textbf{MM-Dist$\downarrow$} & \textbf{Multimodality$\uparrow$}\\ \cmidrule(lr){4-6}
 &  &  & \textbf{Top1} & \textbf{Top2} & \textbf{Top3} & & & \\ \midrule
\multicolumn{9}{c}{\textbf{HumanML3D}} \\ \midrule
Ground Truth   &  - & $0.002^{\pm0.000}$ & $0.511^{\pm0.003}$ & $0.703^{\pm0.003}$ & $0.797^{\pm0.002}$ & $9.503^{\pm0.065}$ & - & -\\ 
MLD \cite{MLD} & CVPR 2023   & $0.473^{\pm0.013}$ & $0.481^{\pm0.003}$ & $0.673^{\pm0.003}$ & $0.772^{\pm0.002}$ & $9.724^{\pm0.082}$& $3.196^{\pm0.010}$ & $2.413^{\pm0.079}$\\
ReMoDiffuse \cite{remodiffuse} & ICCV 2023 & $0.103^{\pm0.004}$ & $0.510^{\pm0.005}$ & $0.698^{\pm0.006}$ & $0.795^{\pm0.004}$ & $9.018^{\pm0.075}$ & $3.025^{\pm0.008}$ & $1.795^{\pm0.043}$\\
MotionDiffuse \cite{motiondiffuse} & TPAMI 2024 & $0.630^{\pm0.001}$ & $0.491^{\pm0.001}$ & $0.681^{\pm0.001}$ & $0.782^{\pm0.001}$ & $9.410^{\pm0.049}$ & $3.113^{\pm0.001}$ & $1.553^{\pm0.042}$\\
MotionLCM \cite{dai2024motionlcmrealtimecontrollablemotion} & ECCV 2024 & $0.467^{\pm0.012}$ & $0.502^{\pm0.003}$ & $0.701^{\pm0.002}$ & $0.803^{\pm0.002}$ & $9.361^{\pm0.660}$ &$3.012^{\pm0.007}$ &$2.172^{\pm0.082}$\\
T2M-GPT \cite{T2M-GPT} & CVPR 2023 & $0.141^{\pm0.005}$ & $0.492^{\pm0.003}$ & $0.679^{\pm0.002}$ & $0.775^{\pm0.002}$ & $9.722^{\pm0.082}$ & $3.121^{\pm0.009}$ &$1.831^{\pm0.048}$\\
MMM \cite{pinyoanuntapong2024mmmgenerativemaskedmotion}& CVPR 2024 & $0.089^{\pm0.002}$ & $0.515^{\pm0.002}$ & $0.708^{\pm0.002}$ & $0.804^{\pm0.002}$& $9.577^{\pm0.050}$ &$2.926^{\pm0.007}$ &$1.226^{\pm0.035}$\\
MoMask \cite{guo2023momaskgenerativemaskedmodeling}& CVPR 2024 & \textcolor{red}{$0.045^{\pm0.002}$} & $0.521^{\pm0.002}$ & $0.713^{\pm0.002}$ & $0.807^{\pm0.002}$& - & $2.958^{\pm0.008}$ & $1.241^{\pm0.040}$\\
MDM \cite{MDM} & ICLR 2023   & $0.544^{\pm0.044}$ & $0.320^{\pm0.005}$ & $0.498^{\pm0.004}$ & $0.611^{\pm0.007}$ & \textcolor{blue}{$9.559^{\pm0.086}$} &$5.556^{\pm0.027}$ &\textcolor{red}{$2.799^{\pm0.072}$} \\
StableMoFusion \cite{stablemofusion} & ACM MM 2024 & $0.152^{\pm0.004}$ & $0.546^{\pm0.002}$ & $0.742^{\pm0.002}$ & $0.835^{\pm0.002}$ & \textcolor{red}{$9.466^{\pm0.002}$} & \textcolor{blue}{$2.781^{\pm0.011}$} &$1.362^{\pm0.062}$\\ 
\midrule

Free-T2M (Ours) & - & \textcolor{blue}{$0.060^{\pm0.004}$} & \textcolor{red}{$0.555^{\pm0.003}$} & \textcolor{red}{$0.752^{\pm0.003}$} & \textcolor{red}{$0.841^{\pm0.002}$} & $9.710^{\pm0.138}$ & \textcolor{red}{$2.749^{\pm0.011}$} & $1.634^{\pm0.102}$\\ 
\midrule
\multicolumn{9}{c}{\textbf{KIT-ML}} \\ \midrule
Ground Truth & - & $0.031^{\pm0.004}$ & $0.424^{\pm0.005}$ & $0.649^{\pm0.006}$ & $0.779^{\pm0.006}$ & $11.080^{\pm0.097}$ & - & - \\ 
MLD \cite{MLD} & CVPR 2023 & $0.404^{\pm0.027}$ & $0.390^{\pm0.008}$ & $0.609^{\pm0.008}$ & $0.734^{\pm0.007}$ & $10.800^{\pm0.117}$ & $3.204^{\pm0.027}$ & \textcolor{blue}{$2.192^{\pm0.071}$}\\
ReMoDiffuse \cite{remodiffuse} & ICCV 2023 & $0.155^{\pm0.006}$ & $0.427^{\pm0.014}$ & $0.641^{\pm0.004}$ & $0.765^{\pm0.055}$ & $10.800^{\pm0.105}$ & \textcolor{red}{$1.239^{\pm0.028}$} & $1.239^{\pm0.028}$ \\
MotionDiffuse \cite{motiondiffuse} & TPAMI 2024  & $1.954^{\pm0.062}$ & $0.417^{\pm0.004}$ & $0.621^{\pm0.004}$ & $0.739^{\pm0.004}$ & \textcolor{blue}{$11.100^{\pm0.143}$} & $2.958^{\pm0.005}$ & $0.730^{\pm0.013}$ \\
T2M-GPT \cite{T2M-GPT} & CVPR 2023 & $0.514^{\pm0.029}$ & $0.416^{\pm0.006}$ & $0.627^{\pm0.006}$ & $0.745^{\pm0.006}$ & $10.921^{\pm0.108}$ & $3.007^{\pm0.023}$ & $1.570^{\pm0.039}$ \\
MotionGPT \cite{motiongpth} & NeurIPS 2023 & $0.510^{\pm0.016}$ & $0.366^{\pm0.005}$ & $0.558^{\pm0.004}$ & $0.680^{\pm0.005}$ & $10.350^{\pm0.084}$ & $3.527^{\pm0.021}$ & \textcolor{red}{$2.328^{\pm0.117}$} \\
MMM \cite{pinyoanuntapong2024mmmgenerativemaskedmotion}& CVPR 2024& $0.316^{\pm0.028}$ & $0.404^{\pm0.005}$ & $0.621^{\pm0.005}$ & $0.744^{\pm0.004}$ & $10.910^{\pm0.101}$ & $2.977^{\pm0.019}$ & $1.232^{\pm0.039}$ \\
MoMask \cite{guo2023momaskgenerativemaskedmodeling}& CVPR 2024 & \textcolor{blue}{$0.204^{\pm0.011}$} & $0.433^{\pm0.007}$ & $0.656^{\pm0.005}$ & $0.781^{\pm0.005}$ & - & $2.779^{\pm0.022}$ & $1.131^{\pm0.043}$ \\
MDM \cite{MDM} & ICLR 2023 & $0.497^{\pm0.021}$ & $0.164^{\pm0.004}$ & $0.291^{\pm0.004}$ & $0.396^{\pm0.004}$ & $10.847^{\pm0.119}$ & $9.191^{\pm0.022}$ & $1.907^{\pm0.214}$ \\
StableMoFusion \cite{stablemofusion} & ACM MM 2024 & $0.258^{\pm0.029}$ & \textcolor{blue}{$0.445^{\pm0.006}$} & \textcolor{blue}{$0.660^{\pm0.005}$} & \textcolor{blue}{$0.782^{\pm0.004}$} & $10.936^{\pm0.077}$ & $2.800^{\pm0.018}$ & $1.362^{\pm0.062}$ \\ \midrule
Free-T2M (Ours) & - & \textcolor{red}{$0.150^{\pm0.019}$} & \textcolor{red}{$0.454^{\pm0.007}$} & \textcolor{red}{$0.675^{\pm0.005}$} & \textcolor{red}{$0.798^{\pm0.007}$} & \textcolor{red}{$10.977^{\pm0.102}$} & \textcolor{blue}{$2.689^{\pm0.020}$} & $1.141^{\pm0.064}$ \\
\bottomrule
\end{tabular}%
}
\vspace{0.1cm}
\caption{Quantitative comparison on the HumanML3D and KIT-ML datasets. $\pm$ indicates a 95\% confidence interval. $\downarrow$: Lower is better. $\uparrow$: Higher is better. $\rightarrow$: Closer to the Ground Truth (GT) is better. \textcolor{red}{Red} and \textcolor{blue}{Blue} indicate the best and the second-best results respectively across all methods for each metric. Our method, Free-T2M, demonstrates state-of-the-art or highly competitive performance across multiple key metrics on both datasets.}
\label{tab:combined_results}
\end{table*}

%% file: tab/ablation_Loss.tex
\begin{table}[t]
\centering
\large
\renewcommand{\arraystretch}{1.5}
\setlength{\belowcaptionskip}{20pt}
\resizebox{\linewidth}{!}{%
\begin{tabular}{lcccc}
\hline
\textbf{Method} & \textbf{FID $\downarrow$} & \multicolumn{3}{c}{\textbf{R-Precision $\uparrow$}}  \\ \cline{3-5}
 &  & \textbf{top1} & \textbf{top2} & \textbf{top3} \\ \hline

Free-T2M (w FTA $f_{full}$)    
                              & \textcolor{red}{$\mathbf{0.060^{\pm0.004}}$} & \textcolor{red}{$\mathbf{0.555^{\pm0.003}}$} & \textcolor{red}{$\mathbf{0.752^{\pm0.003}}$} & \textcolor{red}{$\mathbf{0.841^{\pm0.002}}$}  \\ 

Free-T2M (w FTA $f_l$)    
                              & $0.071^{\pm0.004}$ & $0.552^{\pm0.004}$ & $0.747^{\pm0.004}$ & $0.839^{\pm0.003}$  \\

Free-T2M (w FTA $f_h$)    
                              & $0.064^{\pm0.002}$ & \textcolor{red}{$\mathbf{0.555^{\pm0.004}}$} & $0.748^{\pm0.004}$ & $0.840^{\pm0.003}$  \\ 
Free-T2M (w FTA $x_0$)    
                              & $0.078^{\pm0.004}$ & $0.548^{\pm0.002}$ & $0.744^{\pm0.003}$ & $0.835^{\pm0.002}$  \\

Free-T2M (w/o FTA $f_{full}$)    
                              & $0.129^{\pm0.007}$ & $0.545^{\pm0.003}$ & $0.735^{\pm0.005}$ & $0.823^{\pm0.003}$  \\
Free-T2M (w/o FTA $f_{l}$)    
                              & $0.074^{\pm0.004}$ & $0.546^{\pm0.003}$ & $0.739^{\pm0.003}$ & $0.830^{\pm0.005}$  \\                                
                           
Free-T2M (w/o FTA $f_{h}$)    
                              & $0.108^{\pm0.006}$ & $0.544^{\pm0.003}$ & $0.734^{\pm0.003}$ & $0.826^{\pm0.003}$  \\

                              \hline
                              \rule{0pt}{5mm}
\end{tabular}%
}
\caption{Ablation experiment results for FTA and selection of frequency bands on the HumanML3D test set. Red indicates the best result.}
\label{table:ablation_results}

\end{table}

%% file: sec/5_conclusion.tex
\section{Conclusion}
In this paper, we propose Free-T2M, a novel framework that systematically analyzes the denoising mechanism in motion sequence generation and integrates stage-specific frequency-domain consistency alignment to optimize the denoising process. To the best of our knowledge, this work represents the first comprehensive investigation of the denoising dynamics in motion generation, offering new perspectives and laying a foundation for future research in this field. By explicitly reinforcing semantic consistency during the early denoising stages and enhancing fine-grained detail modeling in the later stages, Free-T2M achieves superior robustness and accuracy across diverse baselines, thereby establishing new state-of-the-art performance within diffusion-based architectures.

%% file: main.bbl
\begin{thebibliography}{10}

\bibitem{MDM}
G.~Tevet, S.~Raab, B.~Gordon, Y.~Shafir, D.~Cohen-Or, and A.~H. Bermano, ``Human motion diffusion model,'' 2022.

\bibitem{SATO}
W.~chen, H.~Xiao, E.~Zhang, L.~Hu, L.~Wang, M.~Liu, and C.~Chen, ``Sato: Stable text-to-motion framework,'' in {\em Proceedings of the 32nd ACM International Conference on Multimedia}, MM ’24, p.~6989–6997, ACM, Oct. 2024.

\bibitem{jiang2024harmonwholebodymotiongeneration}
Z.~Jiang, Y.~Xie, J.~Li, Y.~Yuan, Y.~Zhu, and Y.~Zhu, ``Harmon: Whole-body motion generation of humanoid robots from language descriptions,'' 2024.

\bibitem{luo2024universalhumanoidmotionrepresentations}
Z.~Luo, J.~Cao, J.~Merel, A.~Winkler, J.~Huang, K.~Kitani, and W.~Xu, ``Universal humanoid motion representations for physics-based control,'' 2024.

\bibitem{DDPM}
J.~Ho, A.~Jain, and P.~Abbeel, ``Denoising diffusion probabilistic models,'' 2020.

\bibitem{stablevideodiffusionscaling}
A.~Blattmann, T.~Dockhorn, S.~Kulal, D.~Mendelevitch, M.~Kilian, D.~Lorenz, Y.~Levi, Z.~English, V.~Voleti, A.~Letts, V.~Jampani, and R.~Rombach, ``Stable video diffusion: Scaling latent video diffusion models to large datasets,'' 2023.

\bibitem{ning2025dctdiffintriguingpropertiesimage}
M.~Ning, M.~Li, J.~Su, H.~Jia, L.~Liu, M.~Beneš, W.~Chen, A.~A. Salah, and I.~O. Ertugrul, ``Dctdiff: Intriguing properties of image generative modeling in the dct space,'' 2025.

\bibitem{MLD}
X.~Chen, B.~Jiang, W.~Liu, Z.~Huang, B.~Fu, T.~Chen, J.~Yu, and G.~Yu, ``Executing your commands via motion diffusion in latent space,'' 2023.

\bibitem{stablemofusion}
Y.~Huang, H.~Yang, C.~Luo, Y.~Wang, S.~Xu, Z.~Zhang, M.~Zhang, and J.~Peng, ``Stablemofusion: Towards robust and efficient diffusion-based motion generation framework,'' 2024.

\bibitem{dai2024motionlcmrealtimecontrollablemotion}
W.~Dai, L.-H. Chen, J.~Wang, J.~Liu, B.~Dai, and Y.~Tang, ``Motionlcm: Real-time controllable motion generation via latent consistency model,'' 2024.

\bibitem{T2M-GPT}
J.~Zhang, Y.~Zhang, X.~Cun, S.~Huang, Y.~Zhang, H.~Zhao, H.~Lu, and X.~Shen, ``T2m-gpt: Generating human motion from textual descriptions with discrete representations,'' 2023.

\bibitem{humanmotion3d}
Z.~Cen, H.~Pi, S.~Peng, Z.~Shen, M.~Yang, S.~Zhu, H.~Bao, and X.~Zhou, ``Generating human motion in 3d scenes from text descriptions,'' 2024.

\bibitem{motionx}
J.~Lin, A.~Zeng, S.~Lu, Y.~Cai, R.~Zhang, H.~Wang, and L.~Zhang, ``Motion-x: A large-scale 3d expressive whole-body human motion dataset,'' 2024.

\bibitem{miao2024autonomousllmenhancedadversarialattack}
H.~Miao, F.~Ma, R.~Quan, K.~Zhan, and Y.~Yang, ``Autonomous llm-enhanced adversarial attack for text-to-motion,'' 2024.

\bibitem{Chen2025ANTAN}
W.~Chen, K.~Yu, H.~Jia, K.~Yuan, B.~Tian, S.~Lai, H.~Xiao, E.~Zhang, L.~Wang, and Y.~Yue, ``Ant: Adaptive neural temporal-aware text-to-motion model,'' {\em ArXiv}, vol.~abs/2506.02452, 2025.

\bibitem{firoozi2025foundation}
R.~Firoozi, J.~Tucker, S.~Tian, A.~Majumdar, J.~Sun, W.~Liu, Y.~Zhu, S.~Song, A.~Kapoor, K.~Hausman, {\em et~al.}, ``Foundation models in robotics: Applications, challenges, and the future,'' {\em The International Journal of Robotics Research}, vol.~44, no.~5, pp.~701--739, 2025.

\bibitem{chen2024autotamp}
Y.~Chen, J.~Arkin, C.~Dawson, Y.~Zhang, N.~Roy, and C.~Fan, ``Autotamp: Autoregressive task and motion planning with llms as translators and checkers,'' in {\em 2024 IEEE International conference on robotics and automation (ICRA)}, pp.~6695--6702, IEEE, 2024.

\bibitem{kaelbling2011hierarchical}
L.~P. Kaelbling and T.~Lozano-P{\'e}rez, ``Hierarchical task and motion planning in the now,'' in {\em 2011 IEEE international conference on robotics and automation}, pp.~1470--1477, IEEE, 2011.

\bibitem{vu2024coast}
B.~Vu, T.~Migimatsu, and J.~Bohg, ``Coast: Constraints and streams for task and motion planning,'' in {\em 2024 IEEE International Conference on Robotics and Automation (ICRA)}, pp.~14875--14881, IEEE, 2024.

\bibitem{gatt2018survey}
A.~Gatt and E.~Krahmer, ``Survey of the state of the art in natural language generation: Core tasks, applications and evaluation,'' {\em Journal of Artificial Intelligence Research}, vol.~61, pp.~65--170, 2018.

\bibitem{cohen2024survey}
V.~Cohen, J.~X. Liu, R.~Mooney, S.~Tellex, and D.~Watkins, ``A survey of robotic language grounding: tradeoffs between symbols and embeddings,'' in {\em Proceedings of the Thirty-Third International Joint Conference on Artificial Intelligence}, pp.~7999--8009, 2024.

\bibitem{schaal1996learning}
S.~Schaal, ``Learning from demonstration,'' {\em Advances in neural information processing systems}, vol.~9, 1996.

\bibitem{atkeson1997robot}
C.~G. Atkeson and S.~Schaal, ``Robot learning from demonstration,'' in {\em ICML}, vol.~97, pp.~12--20, 1997.

\bibitem{brys2015reinforcement}
T.~Brys, A.~Harutyunyan, H.~B. Suay, S.~Chernova, M.~E. Taylor, and A.~Now{\'e}, ``Reinforcement learning from demonstration through shaping.,'' in {\em IJCAI}, pp.~3352--3358, 2015.

\bibitem{dubois2023alpacafarm}
Y.~Dubois, C.~X. Li, R.~Taori, T.~Zhang, I.~Gulrajani, J.~Ba, C.~Guestrin, P.~S. Liang, and T.~B. Hashimoto, ``Alpacafarm: A simulation framework for methods that learn from human feedback,'' {\em Advances in Neural Information Processing Systems}, vol.~36, pp.~30039--30069, 2023.

\bibitem{barreiros2025motor}
J.~Barreiros and P.~Passos, {\em Motor Behavior: Control, Learning and Development}.
\newblock Taylor \& Francis, 2025.

\bibitem{huang2024surface}
Z.~Huang, Y.~Wen, Z.~Wang, J.~Ren, and K.~Jia, ``Surface reconstruction from point clouds: A survey and a benchmark,'' {\em IEEE transactions on pattern analysis and machine intelligence}, 2024.

\bibitem{remodiffuse}
M.~Zhang, X.~Guo, L.~Pan, Z.~Cai, F.~Hong, H.~Li, L.~Yang, and Z.~Liu, ``Remodiffuse: Retrieval-augmented motion diffusion model,'' 2023.

\bibitem{motiondiffuse}
M.~Zhang, Z.~Cai, L.~Pan, F.~Hong, X.~Guo, L.~Yang, and Z.~Liu, ``Motiondiffuse: Text-driven human motion generation with diffusion model,'' 2022.

\bibitem{physdiff}
Y.~Yuan, J.~Song, U.~Iqbal, A.~Vahdat, and J.~Kautz, ``Physdiff: Physics-guided human motion diffusion model,'' 2023.

\bibitem{amass}
N.~Mahmood, N.~Ghorbani, N.~F. Troje, G.~Pons-Moll, and M.~J. Black, ``Amass: Archive of motion capture as surface shapes,'' 2019.

\bibitem{li2024infinitemotionextendedmotion}
M.~Li, C.~Zhai, S.~Yao, Z.~Xie, K.~Chen, and Y.-G. Jiang, ``Infinite motion: Extended motion generation via long text instructions,'' 2024.

\bibitem{hu2024efficienttextdrivenmotiongeneration}
M.~Hu, M.~Zhu, X.~Zhou, Q.~Yan, S.~Li, C.~Liu, and Q.~Chen, ``Efficient text-driven motion generation via latent consistency training,'' 2024.

\bibitem{boostingdiffusionmodelsmoving}
Y.~Qian, Q.~Cai, Y.~Pan, Y.~Li, T.~Yao, Q.~Sun, and T.~Mei, ``Boosting diffusion models with moving average sampling in frequency domain,'' 2024.

\bibitem{pinyoanuntapong2024mmmgenerativemaskedmotion}
E.~Pinyoanuntapong, P.~Wang, M.~Lee, and C.~Chen, ``Mmm: Generative masked motion model,'' 2024.

\bibitem{guo2023momaskgenerativemaskedmodeling}
C.~Guo, Y.~Mu, M.~G. Javed, S.~Wang, and L.~Cheng, ``Momask: Generative masked modeling of 3d human motions,'' 2023.

\bibitem{motiongpth}
B.~Jiang, X.~Chen, W.~Liu, J.~Yu, G.~Yu, and T.~Chen, ``Motiongpt: Human motion as a foreign language,'' 2023.

\bibitem{Guo_2022_CVPR}
C.~Guo, S.~Zou, X.~Zuo, S.~Wang, W.~Ji, X.~Li, and L.~Cheng, ``Generating diverse and natural 3d human motions from text,'' in {\em Proceedings of the IEEE/CVF Conference on Computer Vision and Pattern Recognition (CVPR)}, pp.~5152--5161, June 2022.

\bibitem{KIT}
M.~Plappert, C.~Mandery, and T.~Asfour, ``The kit motion-language dataset,'' {\em Big Data}, vol.~4, p.~236–252, Dec. 2016.

\bibitem{AMASS:ICCV:2019}
N.~Mahmood, N.~Ghorbani, N.~F. Troje, G.~Pons-Moll, and M.~J. Black, ``{AMASS}: Archive of motion capture as surface shapes,'' in {\em International Conference on Computer Vision}, pp.~5442--5451, Oct. 2019.

\bibitem{guo2020action2motion}
C.~Guo, X.~Zuo, S.~Wang, S.~Zou, Q.~Sun, A.~Deng, M.~Gong, and L.~Cheng, ``Action2motion: Conditioned generation of 3d human motions,'' in {\em Proceedings of the 28th ACM International Conference on Multimedia}, pp.~2021--2029, 2020.

\bibitem{NIPS2017_8a1d6947}
M.~Heusel, H.~Ramsauer, T.~Unterthiner, B.~Nessler, and S.~Hochreiter, ``Gans trained by a two time-scale update rule converge to a local nash equilibrium,'' in {\em Advances in Neural Information Processing Systems} (I.~Guyon, U.~V. Luxburg, S.~Bengio, H.~Wallach, R.~Fergus, S.~Vishwanathan, and R.~Garnett, eds.), vol.~30, Curran Associates, Inc., 2017.

\end{thebibliography}
